%% file: main.tex
\documentclass[letterpaper]{article} 
\usepackage{aaai2027} 
\nocopyright 
\usepackage[hyphens]{url} 
\usepackage{graphicx} 
\usepackage{natbib} 
\usepackage{caption} 
\usepackage{amsmath}
\usepackage{amssymb}
\usepackage{booktabs}
\usepackage{multirow}
\usepackage{siunitx}
\usepackage{makecell}

\usepackage{algorithm}
\usepackage{algorithmic}

\usepackage{newfloat}
\usepackage{listings}
\DeclareCaptionStyle{ruled}{labelfont=normalfont,labelsep=colon,strut=off} 
\floatstyle{ruled}
\newfloat{listing}{tb}{lst}{}
\floatname{listing}{Listing}

\usepackage[hidelinks]{hyperref}
\hypersetup{
    pdftitle={Long-Tailed 3D Point Cloud Dataset Distillation},
    pdfauthor={Jiahao You, Xu Han, Jinfeng Xu, Xianzhi Li}
}

\title{Long-Tailed 3D Point Cloud Dataset Distillation}
\author{
    Jiahao You\textsuperscript{1}\equalcontrib,
    Xu Han\textsuperscript{1}\equalcontrib,
    Jinfeng Xu\textsuperscript{2},
    Xianzhi Li\textsuperscript{1}\corresponding
}
\affiliations{
    \textsuperscript{1}Huazhong University of Science and Technology\\
    \textsuperscript{2}Hunan University\\
    yuguchenyu@hust.edu.cn,
    xhanxu@hust.edu.cn,
    jinfengxu@hnu.edu.cn,
    xzli@hust.edu.cn
}

\begin{document}

\maketitle

\input{00_abs}

\input{01_intro}

\input{02_related_work}

\input{03_method_problem}

\input{04_method_main}

\input{05_experiment_setup}

\input{06_main_result}

\input{07_ablation_study}

\input{08_conclusion}

\bibliography{references}

\end{document}

%% file: 00_abs.tex
\begin{abstract}

\ifnum1=0
Dataset distillation compresses large-scale datasets into compact synthetic sets while preserving training utility.
Existing point cloud dataset distillation methods mainly focus on permutation invariance and rotation variation, while largely overlooking the long-tailed class distributions commonly observed in point cloud datasets.
Different from conventional long-tailed dataset distillation in the image domain, where the training set is long-tailed but the test set is usually class-balanced, point cloud datasets may be long-tailed in both training and evaluation splits.
In this setting, simply emphasizing tail classes may reduce the utility of head classes, which still dominate the evaluation distribution, making it insufficient to directly apply existing long-tailed distillation strategies. To this end, we propose a long-tailed point cloud dataset distillation framework that exploits long-tail characteristics beyond geometry-aware optimization. Adaptive Synthetic Budgeting assigns class-wise synthetic budgets according to class quantity and distillation difficulty. Given the allocated budget, 3D Long-Tailed Distribution Matching optimizes synthetic point clouds through Global-Local Feature Alignment and Prior-Balanced Calibration, preserving intra-class support while calibrating expert supervision under imbalanced class priors. 
Experiments on multiple point cloud benchmarks show consistent gains over existing dataset distillation, point cloud distillation, and long-tailed distillation methods, including a 7\% improvement over the state of the art on ShapeNet.
\fi

\ifnum1=0
Dataset distillation compresses large-scale datasets into compact synthetic sets while preserving training utility.
Existing point cloud dataset distillation methods mainly adapt this paradigm to 3D data by leveraging point-cloud-specific properties, such as permutation invariance and rotation variation.
Yet real point cloud datasets are not only geometrically structured, but also distributionally imbalanced, with both training and test splits often exhibiting long-tailed class distributions.
This differs from conventional image-domain long-tailed distillation, where the training set is long-tailed but the test set is usually class-balanced.
In long-tailed point cloud test settings, directly applying tail-oriented rebalancing strategies may compromise head-class utility, since head classes can still account for a large portion of the test distribution.
To address this problem, we propose a long-tailed point cloud dataset distillation framework that exploits long-tail characteristics beyond geometry-aware optimization.
Adaptive Synthetic Budgeting assigns class-wise synthetic budgets according to class quantity and distillation difficulty.
Given the allocated budget, 3D Long-Tailed Distribution Matching optimizes synthetic point clouds through Global-Local Feature Alignment and Prior-Balanced Calibration, preserving intra-class support and calibrating expert supervision under imbalanced class priors.
Extensive experiments demonstrate the effectiveness of our framework, achieving 7\% higher accuracy than the state of the art on ShapeNet.
\fi

\ifnum1=1
Dataset distillation compresses large-scale datasets into compact synthetic sets while preserving their training utility,
enabling efficient 3D point cloud training.
Current point cloud dataset distillation methods only tackle geometric and representation challenges while ignoring the distributional imbalance prevalent in point cloud datasets where both training and test splits follow long-tailed class distributions.
To our knowledge, we present the first study on long-tailed point cloud dataset distillation.
Rather than focusing primarily on geometric and representation properties or simply constructing a class-balanced synthetic set, our framework explicitly accounts for long-tailed class distributions via two core modules.
First, we design Adaptive Synthetic Budgeting to allocate class-wise synthetic budgets according to class quantity and the expected benefit of additional synthetic samples.
Given the allocated budgets, we further design 3D Long-Tailed Distribution Matching to optimize synthetic point clouds through Global-Local Feature Alignment and Prior-Aware Supervision.
The former preserves both global class distributions and diverse intra-class structures, while the latter provides class-dependent expert supervision to keep tail-class samples recognizable while maintaining diverse head-class patterns. 
Extensive experiments demonstrate the effectiveness of our method, lifting classification accuracy by 7.0 points on ShapeNet55 against state-of-the-art methods.
\fi

\end{abstract}

%% file: 01_intro.tex
\section{Introduction}
Dataset distillation~\cite{DD} aims to condense a large-scale training set into a small, representative synthetic set while faithfully retaining the training utility of the full original dataset.
By drastically reducing data storage overhead and model training costs, this technique has been widely studied in the image domain as an effective solution to the efficiency bottleneck of model training~\cite{dc,dm,mtt}.
This need is particularly relevant to point cloud learning: since point clouds consist of large sets of 3D points, point cloud datasets often occupy massive storage space and require extremely time-consuming model optimization during training, which hinders the efficient deployment and rapid iteration of deep models.

To bring these efficiency benefits to point cloud learning, dataset distillation has recently been extended to 3D point cloud scenarios, opening a new avenue for efficient 3D visual understanding~\cite{3DDP,sadm,PCC,dd3d}.
Existing methods mainly tailor conventional distillation frameworks to point-cloud geometry and representation, such as unordered point sets~\cite{sadm}, rotation variation, flexible point resolution~\cite{dd3d}, and structural consistency~\cite{3DDP}.
By incorporating techniques such as semantic alignment, rotation-invariant optimization, resolution-flexible synthesis, and structure-aware matching, these methods have made preliminary progress in compact point cloud dataset construction.

\input{fig/motivation}

However, beyond these geometric and representation properties, real-world point cloud datasets also commonly exhibit \textbf{long-tailed class distributions}. 
Due to differences in object occurrence frequency, data collection difficulty, and dataset construction preferences, widely used benchmarks such as ModelNet40, ShapeNet55, ScanObjectNN, and ShapeNetPart show substantial class imbalance, as illustrated in Fig.~\ref{fig:motivation}. 
This important distributional characteristic has been largely overlooked by existing point cloud distillation methods.
They typically follow the standard class-balanced distillation protocol and allocate an equal number of synthetic samples to each class, without adapting the synthetic budget to the long-tailed distribution of the original data.

More importantly, point cloud benchmarks are plagued by a unique \textbf{dual-imbalance condition}, referring to the scenario where class imbalance emerges in both training and test splits simultaneously.
As shown in Fig.~\ref{fig:motivation}, this setup stands apart from conventional image-domain long-tailed dataset distillation: image datasets adopt a long-tailed training set alongside a fully class-balanced test set~\cite{ImageNet-LT,cifar10}.
In stark contrast, 3D point cloud datasets suffer from this dual imbalance on both partitions; for instance, ShapeNet55 achieves an extreme test-set imbalance factor as high as 140.6.
This distinction calls for a different distillation objective.
In image-domain long-tailed dataset distillation, a common goal is to transform a long-tailed training set into a class-balanced synthetic set for balanced recognition~\cite{ltdd,rldd,slltdd}.
Such a formulation is appropriate when the test distribution is class-balanced. 
However, when the test distribution is also long-tailed, enforcing a class-balanced synthetic set under a limited budget may allocate a disproportionately large share of the synthetic budget to tail classes and weaken the representation of head classes, which account for a substantial portion of the test data.
Therefore, \emph{long-tailed point cloud dataset distillation should seek to preserve the utility of the original imbalanced dataset, maintaining strong head-class performance while providing sufficient coverage for tail classes}.

Achieving this objective raises two coupled problems.
(1) \textbf{How to allocate the limited synthetic budget across imbalanced classes?} Assigning the same number of synthetic samples to every class ignores the original long-tailed distribution, whereas simply preserving the original class ratios may not yield the best performance under a limited budget.
(2) \textbf{How to optimize the synthetic point clouds once the class-wise budget is determined?}
Existing distillation objectives, such as distribution matching, are generally designed without explicitly considering long-tailed class distributions. How to adapt them to preserve the utility of synthetic point clouds under class imbalance remains underexplored in point cloud dataset distillation.

To address these two problems, we propose a long-tailed point cloud dataset distillation framework with two new components.
First, Adaptive Synthetic Budgeting (ASB) allocates the synthetic budget across classes by considering both class quantity and the expected benefit of additional synthetic samples.
This allows the class-wise composition of the synthetic set to adapt to the long-tailed class distribution of the original dataset.
Second, given the allocated budget, 3D Long-Tailed Distribution Matching (3D-LTDM) optimizes the synthetic point clouds with a long-tail-aware distribution matching objective.
It includes Global-Local Feature Alignment (GLFA), which combines class-level alignment with local cluster-level alignment to preserve diverse intra-class structures, and Prior-Aware Supervision (PAS), which uses class-prior information to provide class-dependent expert supervision.
In this way, ASB determines how many synthetic samples to allocate to each class, while 3D-LTDM determines how to optimize these samples under long-tailed distributions.

We conduct extensive comparisons on various long-tailed point cloud recognition datasets, and further extend our experiments to segmentation tasks and even image datasets. Consistent results verify the superiority of our method and demonstrate the necessity of accounting for long-tailed class distributions.
%
%
Our contributions are summarized as follows:
\begin{itemize}
\item We identify long-tailed class distributions as an overlooked challenge in point cloud dataset distillation and study a realistic setting where both the training and test sets can exhibit long-tailed distributions.
\item We propose Adaptive Synthetic Budgeting (ASB), which allocates the synthetic budget across classes according to class quantity and the expected benefit of additional synthetic samples.
\item We propose 3D Long-Tailed Distribution Matching (3D-LTDM), which combines Global-Local Feature Alignment and Prior-Aware Supervision to preserve diverse intra-class structures and provide class-dependent expert supervision under long-tailed distributions.
\end{itemize}

%% file: fig/motivation.tex
\begin{figure}[t]
    \centering
    \includegraphics[width=\linewidth]{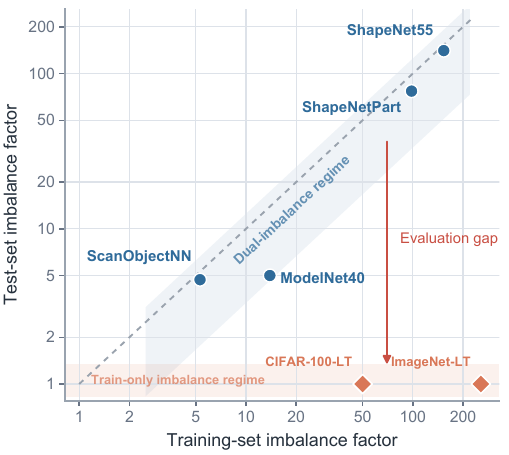}
    \caption{
    Train–test class imbalance across benchmarks. The imbalance factor (IF) is the ratio of the largest to the smallest class size. Conventional 2D image long-tailed settings retain balanced test sets, unlike the dual imbalance in 3D point cloud datasets.
    }
    \label{fig:motivation}
    \vspace{-10pt}
\end{figure}

%% file: 02_related_work.tex
\section{Related Work}

\input{fig/method_framework}

\subsection{Image Dataset Distillation}

Image dataset distillation has been developed through several major optimization paradigms.
Gradient matching aligns the parameter updates induced by the original and synthetic data~\cite{dc}, while trajectory matching seeks to reproduce the optimization trajectories obtained from training on the original data~\cite{mtt}. Distribution matching offers a more efficient alternative by aligning the feature distributions of the original and synthetic data~\cite{dm}.
More recent methods further improve synthetic data quality through soft labels, parameterizations, and generative models~\cite{sre2l,edc,datm,ptqdc,diffusion}.
All of the above approaches assume balanced class distributions, failing to analyze the influence of class imbalance on synthetic data construction.

To tackle this overlooked problem, recent works extend image dataset distillation to long-tailed scenarios, where the training set exhibits skewed class distributions, while the test set remains class-balanced.
Accordingly, the core paradigm adopted by these methods is to construct class-balanced synthetic sets and mitigate bias derived from long-tailed experts.
For instance, LTDD~\cite{ltdd} reduces the effect of biased expert trajectories through distribution-agnostic matching, while subsequent works introduce unbiased recovery~\cite{rldd} or soft-label correction~\cite{slltdd}.
In contrast, we focus on point cloud benchmarks under a dual-imbalance setting.
Unlike prior methods that eliminate long-tailed distributions and debias expert knowledge, our approach preserves the inherent long-tailed data utility under limited synthetic budgets and adapts supervision via class-prior information.

\subsection{Point Cloud Dataset Distillation}

PCC~\cite{PCC} first introduced dataset distillation for 3D point clouds.
SADM~\cite{sadm} addresses the unordered nature of point clouds and rotation variation through rotation optimization and semantic alignment.
DD3D~\cite{dd3d} enables rotation-invariant distillation and flexible-resolution synthesis.
3DDP~\cite{3DDP} reduces the storage cost of distilled samples through a compact parameterization and preserves structural consistency with a uniformity-aware matching loss.
While these methods consider point-cloud geometry and representation, they do not explicitly account for class distribution in their distillation design.
Our work addresses this gap by examining how long-tailed class distributions affect the utility of the distilled set.




%% file: fig/method_framework.tex

\begin{figure*}[t] 
\centering 
\includegraphics[width=\textwidth]{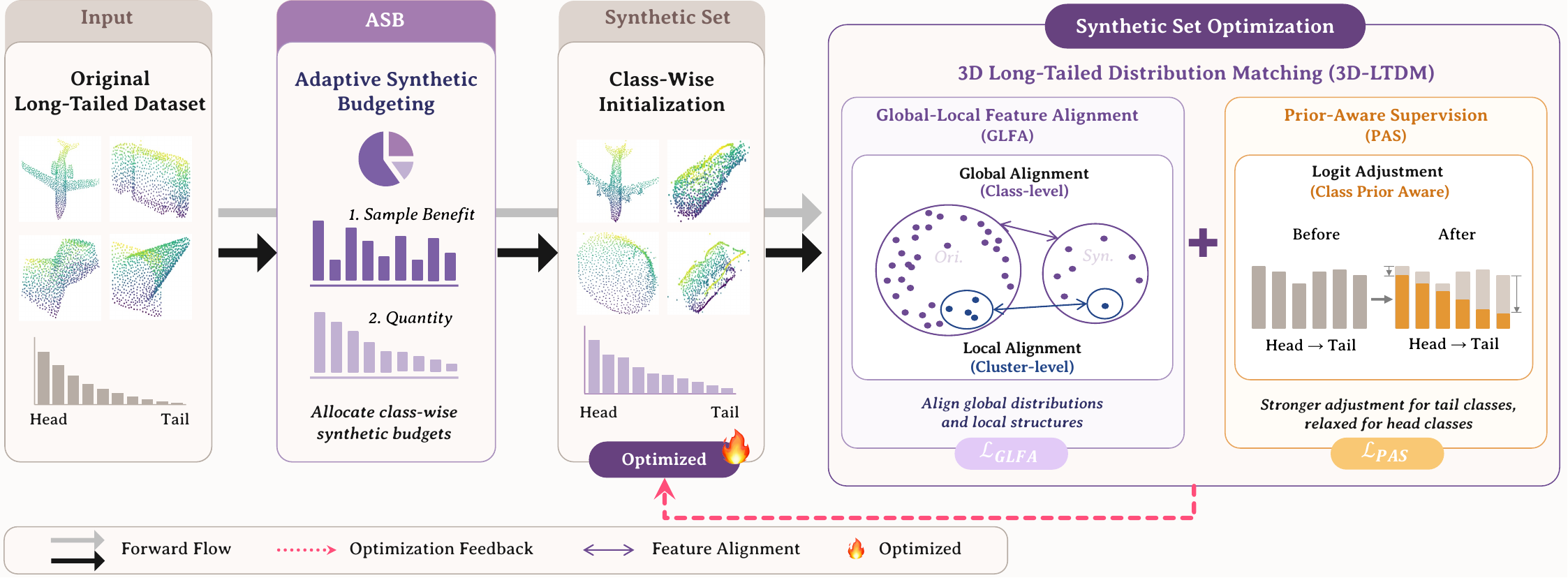} 
\caption{Overview of our framework. Given an original long-tailed dataset and a fixed total synthetic budget, ASB adaptively allocates the budget across classes. The resulting synthetic point clouds are then optimized by 3D-LTDM, where GLFA aligns global and local feature distributions, while PAS applies prior-aware logit adjustment.} 
\label{fig:framework} 
\vspace{-10pt}
\end{figure*}

%% file: 03_method_problem.tex
\section{Methodology}

\subsection{Problem Formulation}


Given a long-tailed point cloud training set
$D_{\mathrm{ori}}=\{D_{\mathrm{ori}}^c\}_{c=1}^{C}$,
where $D_{\mathrm{ori}}^c$ denotes the samples of class $c$ and
$n_c=|D_{\mathrm{ori}}^c|$ is the class size, we order the classes by
sample frequency without loss of generality:
\begin{equation}
n_1 \ge n_2 \ge \cdots \ge n_C,
\qquad
\gamma_{\mathrm{ori}}=\frac{n_1}{n_C}.
\label{eq:original_imbalance_factor}
\end{equation}
Following \citet{classbalancedloss}, $\gamma_{\mathrm{ori}}$ is defined as
the imbalance factor of the original training set, where
$\gamma_{\mathrm{ori}}=1$ indicates a balanced dataset and larger values
indicate more severe class imbalance.

Long-tailed point cloud dataset distillation aims to learn a compact
synthetic set $D_{\mathrm{syn}}=\{D_{\mathrm{syn}}^c\}_{c=1}^{C}$ from
$D_{\mathrm{ori}}$.
Let $m_c=|D_{\mathrm{syn}}^c|$ denote the number of synthetic samples
assigned to class $c$, and let
$M=|D_{\mathrm{syn}}|=\sum_{c=1}^{C}m_c$ denote the total synthetic budget,
where $M \ll |D_{\mathrm{ori}}|$.
Let $\theta_{\mathrm{ori}}$ and $\theta_{\mathrm{syn}}$ denote the models
trained on $D_{\mathrm{ori}}$ and $D_{\mathrm{syn}}$, respectively.
The objective is to minimize the loss discrepancy between the two models:
\begin{equation}
D_{\mathrm{syn}}^{\star}
=
\arg\min_{D_{\mathrm{syn}}}
\left|
\mathbb{E}_{(\mathbf{x},y)\sim\mathcal{P}}
\left[
\ell(f_{\theta_{\mathrm{syn}}}(\mathbf{x}),y)
-
\ell(f_{\theta_{\mathrm{ori}}}(\mathbf{x}),y)
\right]
\right|,
\label{eq:dd_objective}
\end{equation}
where $\mathbf{x}$ is a point cloud, $y$ is its label,
$f_{\theta}$ denotes the prediction model parameterized by $\theta$,
$\ell(\cdot,\cdot)$ is the task loss, and $\mathcal{P}$ represents the
evaluation distribution.


Unlike image-domain long-tailed dataset distillation~\cite{ltdd,rldd}, which often distills a long-tailed training set into a class-balanced synthetic set for balanced recognition, our setting does not impose the constraint $m_1=m_2=\cdots=m_C$. Instead, the class-wise synthetic budget is allowed to vary across classes under the fixed total budget $M$, with the goal of improving the test performance of the model trained on $D_{\mathrm{syn}}$.

%% file: 04_method_main.tex
\subsection{Overall Framework}
As illustrated in Fig.~\ref{fig:framework}, our framework follows a budget-then-optimize pipeline. Given an imbalanced $D_{\mathrm{ori}}$, Adaptive Synthetic Budgeting (ASB) first allocates a dedicated synthetic budget across classes. The distilled point clouds $D_{\mathrm{syn}}$ are then initialized with representative original samples selected around K-means centroids~\cite{kmeans}, and further optimized by 3D Long-Tailed Distribution Matching (3D-LTDM). The following subsections elaborate on ASB and 3D-LTDM.

\subsection{Adaptive Synthetic Budgeting}



Since our setting does not enforce a class-balanced synthetic set, we first determine the class-wise synthetic budget $\{m_c\}_{c=1}^{C}$ before distillation, where $m_c$ denotes the final number of synthetic samples assigned to class $c$. 
Existing methods typically use a uniform budget across classes.
However, under long-tailed distributions, different classes may require different budgets due to their different sample quantities and sensitivities to the available budget. 
We therefore introduce Adaptive Synthetic Budgeting (ASB) to allocate the total budget $M$ across classes.



For each class $c$, we estimate a sample benefit score $g_c$ by comparing
the class-wise validation accuracy $a_c^{\mathrm{exp}}$ of the expert with
the average class-wise validation accuracy $\bar{a}_c$ of proxy models
trained on small class-balanced subsets:
\begin{equation}
g_c
=
\max\!\left(a_c^{\mathrm{exp}}-\bar{a}_c, 0\right)
+
\epsilon,
\label{eq:asb_gain}
\end{equation}
where both accuracies are measured on a held-out validation split
constructed from the official training set, and $\epsilon$ is a small
smoothing constant. A larger gap indicates that class $c$ suffers more
under a limited training budget and may benefit more from additional
synthetic samples.


We combine the sample benefit score with the class size using square-root scaling to obtain the allocation weight:
\begin{equation}
w_c
=
\left(g_c\sqrt{n_c}\right)^{\alpha},
\label{eq:asb_weight}
\end{equation}
where $\alpha$ controls the strength of adaptive allocation. The square-root scaling balances class quantity and prevents head classes from dominating the budget.

To ensure that every class receives a minimum number of synthetic samples, we first reserve $m_{\min}$ samples for each class and allocate the remaining budget according to $w_c$:
\begin{equation}
m_c
=
m_{\min}
+
\left(M-Cm_{\min}\right)
\frac{w_c}{\sum_{j=1}^{C}w_j},
\label{eq:asb_budget}
\end{equation}
where $M\ge C m_{\min}$. We use largest-remainder rounding to obtain integer budgets while preserving $\sum_{c=1}^{C}m_c=M$. ASB only determines the class-wise synthetic composition; the synthetic samples are optimized by the 3D long-tailed distribution matching objective introduced next.

\subsection{3D Long-Tailed Distribution Matching}

\paragraph{From Standard DM to Long-Tailed DM.}
The next key problem is how to optimize the initialized synthetic dataset built under budget constraints to align with the performance of the original dataset.
Following existing dataset distillation methods~\cite{cafe,dance,idm}, we build on a Distribution Matching objective that combines feature distribution matching with an expert classification loss:
\begin{equation}
\mathcal{L}_{\mathrm{DM}}
=
\mathcal{L}_{\mathrm{dm}}
+
\lambda_{\mathrm{cls}}\mathcal{L}_{\mathrm{cls}}.
\label{eq:standard_dm_loss}
\end{equation}

Given the original and synthetic samples of class $c$, denoted as $D_{\mathrm{ori}}^c$ and $D_{\mathrm{syn}}^c$, the feature matching loss is commonly defined as:
\begin{equation}
\mathcal{L}_{\mathrm{dm}}
=
\sum_{c=1}^{C}
\operatorname{dis}\!\left(
\phi_{\mathrm{mid}}(D_{\mathrm{ori}}^c),
\phi_{\mathrm{mid}}(D_{\mathrm{syn}}^c)
\right),
\label{eq:standard_dm}
\end{equation}
where $\phi_{\mathrm{mid}}$ denotes the intermediate feature extractor and $\operatorname{dis}(\cdot,\cdot)$ measures the discrepancy between two feature distributions. Following previous work~\cite{sadm}, we instantiate $\operatorname{dis}(\cdot,\cdot)$ as M3D~\cite{m3d} in our experiments.
The expert classification loss encourages each synthetic sample to be recognized as its target class by the expert model:
\begin{equation}
\mathcal{L}_{\mathrm{cls}}
=
\frac{1}{M}
\sum_{i=1}^{M}
\ell_{\mathrm{ce}}\!\left(
f_{\theta_e}(x_i^{\mathrm{syn}}),
y_i^{\mathrm{syn}}
\right),
\label{eq:standard_cls}
\end{equation}
where $f_{\theta_e}$ denotes the fixed expert model.

However, standard DM is not fully suited to long-tailed point cloud
distillation. 
Its feature matching treats each class as a single global distribution, which may fail to capture dispersed local structures of tail classes under long-tailed distributions.
Meanwhile, its expert classification term enforces identical supervision over all classes, even though expert knowledge captures uneven proportions of head and tail classes. We resolve both limitations via Global-Local Feature Alignment and Prior-Aware Supervision, respectively.

\input{fig/feature_analysis}

\input{tab/pointnetresult}

\paragraph{Global-Local Feature Alignment.}
Standard feature distribution matching aligns original and synthetic samples at the class level. 
However, long-tailed training can make tail-class features more dispersed than head-class features~\cite{bce3s}. 
As illustrated in Fig.~\ref{fig:feature_distribution}, head-class features form compact clusters, while tail-class features are sparse and scattered. 
In this case, a single global alignment may be unreliable and push synthetic samples into low-density regions. 
We therefore introduce Global-Local Feature Alignment (GLFA), which combines global class-level alignment with local cluster-level alignment.

Specifically, we denote the original and synthetic feature sets of class $c$ as $T_c=\phi_{\mathrm{mid}}(D_{\mathrm{ori}}^c)$ and $S_c=\phi_{\mathrm{mid}}(D_{\mathrm{syn}}^c)$, respectively.
The global alignment term follows standard class-wise distribution matching: 
\begin{equation} 
\mathcal{L}_{\mathrm{global}} 
= 
\sum_{c=1}^{C} 
\operatorname{dis}(T_c,S_c). \label{eq:global_loss} 
\end{equation}

%
To capture local intra-class variations, we extract $T_c$ using the fixed expert and partition it into $K_c$ clusters using K-means before distillation; the cluster assignments remain fixed during optimization. 
We set $K_c=m_c$ and optimize each synthetic sample to align with one cluster. 
Let $T_{c,k}$ denote the original features in the $k$-th cluster and $S_{c,k}$ denote the feature of its corresponding synthetic sample.
The local alignment term is defined as:
\begin{equation}
\mathcal{L}_{\mathrm{local}}
=
\sum_{c=1}^{C}
\sum_{k=1}^{K_c}
\operatorname{dis}(T_{c,k},S_{c,k}).
\label{eq:local_loss}
\end{equation}

The GLFA objective combines global and local alignment:
\begin{equation}
\mathcal{L}_{\mathrm{GLFA}}
=
\mathcal{L}_{\mathrm{global}}
+
\lambda_{\mathrm{local}}\mathcal{L}_{\mathrm{local}}.
\label{eq:glfa_loss}
\end{equation}

\paragraph{Prior-Aware Supervision.}


Under long-tailed training, the expert can correctly recognize only part of the tail-class distribution, while covering a broader range of head-class samples.
During distribution matching, tail-class synthetic samples can therefore more easily drift away from the part of the distribution that the expert correctly recognizes. 
They require stronger classification supervision to remain recognizable by the expert. 
In contrast, overly strong supervision can pull head-class synthetic samples toward a few high-confidence regions, reducing their intra-class diversity. 
We therefore introduce Prior-Aware Supervision (PAS), which strengthens the classification constraint for tail classes while relaxing it for head classes.

Let $\mathbf{z}_i=f_{\theta_e}(x_i^{\mathrm{syn}})\in\mathbb{R}^{C}$ denote the logits produced by the fixed expert model for a synthetic sample $x_i^{\mathrm{syn}}$. 
Given class size $n_c$$=$$|D_{\mathrm{ori}}^c|$, we define the empirical class prior as:
\begin{equation}
\pi_c
=
\frac{n_c}{\sum_{j=1}^{C}n_j},
\label{eq:class_prior}
\end{equation}
and adjust the logits according to the class prior:
\begin{equation}
\tilde{z}_i^{(c)}
=
z_i^{(c)}
+
\tau \log \pi_c,
\label{eq:prior_aware_logit}
\end{equation}
where $\tau$ controls the strength of the adjustment.
%
%
Since $\log \pi_c$ is smaller for tail classes, the corresponding class logit is reduced more strongly. 
Minimizing the adjusted loss therefore encourages tail-class synthetic samples to produce stronger evidence under the original expert logits. Head-class samples receive a smaller adjustment, allowing feature matching to better preserve their intra-class diversity.

Hence, the Prior-Aware Supervision loss is defined as:
\begin{equation}
\mathcal{L}_{\mathrm{PAS}}
=
\frac{1}{M}
\sum_{i=1}^{M}
\ell_{\mathrm{ce}}\!\left(
\tilde{\mathbf{z}}_i,
y_i^{\mathrm{syn}}
\right).
\label{eq:pas_loss}
\end{equation}


Finally, the proposed 3D Long-Tailed Distribution Matching objective is: 
\begin{equation}
\mathcal{L}_{\mathrm{LTDM}}
=
\mathcal{L}_{\mathrm{GLFA}}
+
\lambda_{\mathrm{pas}}\mathcal{L}_{\mathrm{PAS}}.
\label{eq:ltdm_loss}
\end{equation}

%% file: fig/feature_analysis.tex
\begin{figure}[t]
    \centering
    \includegraphics[width=\linewidth]{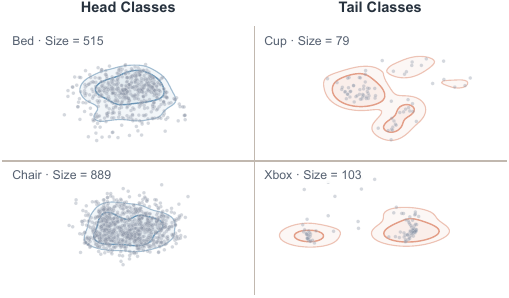}
\caption{
PCA visualization of feature-space distributions on ModelNet40, where tail-class features are sparser.
}
    \label{fig:feature_distribution}
    \vspace{-10pt}
\end{figure}

%% file: tab/pointnetresult.tex
\begin{table*}[t]
    \centering
    \small
    \setlength{\tabcolsep}{1.5pt}
    \begin{tabular*}{\textwidth}{@{\extracolsep{\fill}}l|ccc|ccc|ccc|ccc@{}}
    \toprule
    \multirow{2}{*}{Method}
    & \multicolumn{3}{c|}{ModelNet10}
    & \multicolumn{3}{c|}{ModelNet40}
    & \multicolumn{3}{c|}{ShapeNet55}
    & \multicolumn{3}{c}{ScanObjectNN} \\
    & 1 & 3 & 10
    & 1 & 3 & 10
    & 1 & 3 & 10
    & 1 & 3 & 10 \\
    \midrule
    Random
    & 34.0$\pm$3.6 & 74.9$\pm$2.2 & 84.1$\pm$0.9
    & 32.9$\pm$2.9 & 59.7$\pm$1.1 & 74.5$\pm$0.5
    & 31.8$\pm$1.3 & 59.8$\pm$0.8 & 71.0$\pm$0.4
    & 15.2$\pm$2.1 & 18.9$\pm$1.1 & 34.0$\pm$1.2 \\

    Herding
    & 37.8$\pm$3.9 & 75.6$\pm$1.2 & 86.4$\pm$0.7
    & 52.1$\pm$2.8 & 66.8$\pm$1.5 & 76.4$\pm$0.6
    & 49.5$\pm$2.4 & 63.1$\pm$1.9 & 72.0$\pm$1.2
    & 15.5$\pm$3.0 & 24.6$\pm$1.0 & 36.3$\pm$2.0 \\

    K-Center
    & 37.8$\pm$3.9 & 73.8$\pm$3.4 & 75.3$\pm$1.1
    & 52.1$\pm$2.9 & 56.3$\pm$2.3 & 66.5$\pm$0.6
    & 49.5$\pm$2.4 & 51.8$\pm$1.8 & 50.5$\pm$0.3
    & 15.5$\pm$3.0 & 17.2$\pm$1.4 & 21.3$\pm$1.2 \\
    \midrule

    DM
    & 35.3$\pm$4.6 & 77.7$\pm$1.1 & 85.0$\pm$0.9
    & 52.9$\pm$2.4 & 67.2$\pm$1.0 & 75.5$\pm$0.7
    & 46.8$\pm$2.3 & 63.1$\pm$1.8 & 73.4$\pm$0.5
    & 16.6$\pm$1.8 & 22.6$\pm$2.0 & 37.2$\pm$1.6 \\

    DC
    & 33.9$\pm$4.3 & 74.0$\pm$2.9 & 84.3$\pm$1.1
    & 53.8$\pm$2.7 & 68.0$\pm$1.4 & 77.5$\pm$0.7
    & 48.7$\pm$1.4 & 64.1$\pm$1.0 & 74.4$\pm$0.8
    & 15.1$\pm$2.5 & 25.8$\pm$2.1 & 38.6$\pm$1.6 \\

    MTT
    & 37.3$\pm$3.9 & 73.0$\pm$1.3 & 85.9$\pm$1.3
    & 53.7$\pm$2.9 & 66.2$\pm$1.6 & 74.5$\pm$0.5
    & 50.4$\pm$2.0 & 61.4$\pm$1.4 & 71.0$\pm$0.8
    & 14.2$\pm$3.7 & 27.7$\pm$2.0 & 37.1$\pm$1.5 \\
    \midrule

    PCC
    & 41.1$\pm$3.0 & 78.3$\pm$2.4 & 85.5$\pm$0.6
    & 53.5$\pm$1.5 & 69.5$\pm$1.4 & 78.9$\pm$0.5
    & 51.8$\pm$3.4 & 66.9$\pm$1.6 & 73.4$\pm$0.8
    & 16.4$\pm$2.1 & 24.9$\pm$2.2 & 38.4$\pm$1.3 \\

    SADM
    & \underline{44.7$\pm$6.1} & \underline{84.4$\pm$1.2} & 87.8$\pm$1.0
    & 55.8$\pm$1.5 & 72.1$\pm$0.8 & 80.1$\pm$0.4
    & 54.4$\pm$2.9 & 66.8$\pm$1.3 & 74.7$\pm$0.5
    & \underline{17.4$\pm$1.5} & \underline{31.6$\pm$1.0} & 43.9$\pm$1.9 \\

    LTDD
    & 39.1$\pm$3.5 & 73.3$\pm$1.2 & 87.0$\pm$0.8
    & 54.4$\pm$2.6 & 69.2$\pm$1.4 & 78.5$\pm$0.4
    & 51.8$\pm$1.5 & 62.3$\pm$1.3 & 71.5$\pm$0.4
    & 14.7$\pm$3.7 & 29.3$\pm$1.8 & 38.3$\pm$1.2 \\

    DANCE
    & 38.7$\pm$5.4 & 82.2$\pm$1.5 & 88.3$\pm$0.9
    & 55.3$\pm$1.1 & 70.1$\pm$1.4 & 80.6$\pm$0.6
    & \underline{57.7$\pm$1.4} & 68.3$\pm$1.4 & 75.4$\pm$0.5
    & 15.5$\pm$1.6 & 29.7$\pm$1.7 & 43.0$\pm$1.2 \\

    TGDD
    & 44.2$\pm$3.2 & 82.6$\pm$1.8 & \underline{88.9$\pm$0.5}
    & \underline{56.0$\pm$2.7} & \underline{73.2$\pm$0.7} & \underline{81.2$\pm$0.3}
    & \underline{57.7$\pm$2.3} & \underline{71.3$\pm$1.4} & \underline{76.1$\pm$0.4}
    & 15.3$\pm$2.7 & 30.1$\pm$1.9 & \underline{44.8$\pm$1.4} \\

    \midrule

    Ours
    & \textbf{48.9$\pm$3.8} & \textbf{85.6$\pm$1.3} & \textbf{91.5$\pm$0.5}
    & \textbf{57.2$\pm$1.9} & \textbf{76.7$\pm$0.6} & \textbf{84.2$\pm$0.4}
    & \textbf{59.6$\pm$0.6} & \textbf{76.4$\pm$0.4} & \textbf{83.1$\pm$0.2}
    & \textbf{19.1$\pm$1.5} & \textbf{33.8$\pm$1.3} & \textbf{48.6$\pm$1.2} \\
    \midrule

    Ori. Dataset
    & \multicolumn{3}{c|}{92.4}
    & \multicolumn{3}{c|}{88.3}
    & \multicolumn{3}{c|}{87.5}
    & \multicolumn{3}{c}{65.2} \\
    \bottomrule
    \end{tabular*}
    \caption{
    Classification accuracy (\%) comparing coreset selection and dataset distillation methods across different PPC settings. All methods use the same total synthetic budget, and ``Ori. Dataset'' denotes training on the full original dataset.
    }
    \label{tab:distill_transposed}
    \vspace{-10pt}
\end{table*}

%% file: 05_experiment_setup.tex
\section{Experiments}

\subsection{Experimental Setup}

\subsubsection{Baselines.}
We compare our method with representative coreset selection, general dataset distillation, long-tailed dataset distillation, and point cloud dataset distillation methods. In particular, we include LTDD~\cite{ltdd}, the most closely related long-tailed dataset distillation method, as well as the point cloud distillation methods PCC~\cite{PCC} and SADM~\cite{sadm}.

\subsubsection{Implementation Details.}
Following prior point cloud distillation work, we use PointNet~\cite{pointnet} as the default backbone. To evaluate the intrinsic quality of the distilled samples~\cite{ddranking}, we do not use soft labels or data augmentation in the point cloud experiments. For baseline methods, PPC $=X$ denotes $X$ synthetic point clouds per class. For our method, PPC $=X$ denotes the same total budget of $X \times C$ samples, while ASB adaptively allocates this budget across classes. All results are averaged over five independent runs.

%% file: 06_main_result.tex
\subsection{Main Results}
\subsubsection{Object Classification.}

We evaluate our method on four long-tailed point cloud classification benchmarks: ModelNet10~\cite{modelnet}, ModelNet40~\cite{modelnet}, ShapeNet55~\cite{shapenet2015}, and ScanObjectNN~\cite{scanobjectnn}. We refer to them as MN10, MN40, SN, and SONN, respectively.
%
Table~\ref{tab:distill_transposed} reports the results under different PPC settings. Our method consistently achieves the best performance across all datasets and synthetic budgets, demonstrating the importance of accounting for long-tailed class distributions in point cloud dataset distillation. The gains are particularly pronounced on ShapeNet55, which has the most severe imbalance. Compared with the strongest baseline, our method improves accuracy by $1.9$, $5.1$, and $7.0$ points at PPC $=1$, $3$, and $10$, respectively. Notably, at PPC $=1$, ASB reduces to a uniform one-sample-per-class allocation, yet our method still outperforms all baselines. This result confirms that 3D-LTDM remains effective even when adaptive budget allocation provides no additional flexibility.

\input{tab/part}
\input{tab/corss_architecture}

\input{tab/image_dataset}
\input{fig/feature_compare}

We also observe that SADM and PCC, despite being specifically designed for point cloud dataset distillation, do not consistently outperform image-domain distillation baselines. This suggests that modeling  geometric and representation properties alone may be insufficient when the underlying class distribution is long-tailed. Similarly, directly applying LTDD yields limited improvements in our setting, likely because it is designed for long-tailed training sets with class-balanced test sets, whereas our benchmarks exhibit long-tailed distributions in both splits.

\subsubsection{Part Segmentation.}

Most existing dataset distillation methods, especially those in the image domain, are designed and evaluated primarily for classification, while their effectiveness on dense prediction tasks remains less explored. To evaluate the generality of our framework beyond classification, we extend it to point cloud part segmentation on ShapeNetPart~\cite{shapenet2015}. Specifically, we adopt PointNet as the segmentation backbone, perform distribution matching on its global features, and replace the classification loss with a point-wise segmentation loss. As shown in Table~\ref{tab:segmentation_results}, our method consistently outperforms existing point cloud dataset distillation methods across different PPC settings. Compared with the strongest baseline, it improves mIoU$_\mathrm{I}$ from $61.8$ to $66.7$ at PPC $=3$ and from $68.3$ to $73.3$ at PPC $=10$, corresponding to gains of $4.9$ and $5.0$ points, respectively. These results demonstrate that our distillation strategy remains effective beyond classification and generalizes well to fine-grained point-level prediction.

\subsubsection{Cross-Architecture Generalization.}

To assess transferability beyond the distillation backbone, we train evaluation models using
PointNet++~\cite{pointnet++},
DGCNN~\cite{dgcnn},
PointConv~\cite{wu2018pointconv},
PCT~\cite{pct}, and
PointMLP~\cite{pointmlp}.
As shown in Table~\ref{tab:cross_architecture_results}, our method
consistently outperforms PCC and SADM across
all five evaluation backbones.
The advantage is particularly pronounced on ShapeNet55, the dataset
with the most severe class imbalance, where our method exceeds the
strongest baseline by $7.6$--$10.3$ points.
These results demonstrate the strong cross-architecture generalization
of the distilled point clouds.

\subsubsection{Experiments on Image Datasets.}




To evaluate the transferability beyond point clouds, we conduct experiments on
CIFAR-10-LT and CIFAR-100-LT~\cite{cifar10} and compare our method with
LTDD and RLDD~\cite{rldd}.
Since these benchmarks only provide long-tailed training sets with balanced test sets, we simulate long-tailed evaluation using weighted accuracy based on training class frequencies:
\begin{equation}
\mathrm{WAcc} = \frac{\sum_{c=1}^{C} n_c\,\mathrm{Acc}_c}{\sum_{c=1}^{C} n_c}.
\label{eq:wacc}
\end{equation}
where $n_c$ and $\mathrm{Acc}_c$ denote the training-set size and test accuracy of class $c$, respectively.

Following LTDD and RLDD, we use soft labels during distillation. As shown in Table~\ref{tab:cifar_lt}, our method consistently outperforms both competitors across the two datasets.
These results demonstrate that our approach also remains effective beyond point clouds under a dual-imbalance setting.


\subsubsection{Feature-Space Analysis.}

We further compare our method with TGDD~\cite{tgdd}, the strongest distribution-matching baseline, by visualizing tail-class feature distributions in Fig.~\ref{fig:visual}.
Although ASB tends to allocate fewer synthetic samples to tail classes, our samples better capture their sparse and dispersed structures and more evenly cover the original feature regions.
In contrast, TGDD produces more concentrated samples, resulting in less complete coverage of the tail-class distributions.


%% file: tab/part.tex
\begin{table}[t]
    \centering
    \small
    \setlength{\tabcolsep}{3pt}
    \begin{tabular*}{\columnwidth}{@{\extracolsep{\fill}}clccc@{}}
    \toprule
    PPC & Method & OA (\%) & mIoU$_\mathrm{I}$ (\%) & mIoU$_\mathrm{C}$ (\%) \\
    \midrule
    \multirow{3}{*}{3}
        & PCC  & 68.6$\pm$1.3 & 60.3$\pm$1.1 & 59.5$\pm$0.7 \\
        & SADM & 70.6$\pm$1.1 & 61.8$\pm$0.9 & 59.8$\pm$0.7 \\
        & Ours & \textbf{76.8$\pm$1.9} & \textbf{66.7$\pm$1.0} & \textbf{61.3$\pm$1.2} \\
    \midrule
    \multirow{3}{*}{10}
        & PCC  & 78.1$\pm$0.9 & 68.3$\pm$0.6 & 66.8$\pm$0.9 \\
        & SADM & 78.4$\pm$1.0 & 68.0$\pm$0.9 & 66.1$\pm$0.6 \\
        & Ours & \textbf{85.3$\pm$0.9} & \textbf{73.3$\pm$1.0} & \textbf{68.2$\pm$1.5} \\
    \midrule
    \multicolumn{2}{c}{Original Dataset}
        & 92.7 & 82.4 & 77.1 \\
    \bottomrule
    \end{tabular*}
    \caption{
    Part segmentation on ShapeNetPart. OA, mIoU$_\mathrm{I}$, and mIoU$_\mathrm{C}$ denote overall accuracy, instance-level mean IoU, and class-level mean IoU, respectively. 
    }
    \label{tab:segmentation_results}
\end{table}

%% file: tab/corss_architecture.tex
\begin{table}[t]
    \centering
    \small
    \setlength{\tabcolsep}{2.5pt}
    \renewcommand{\arraystretch}{1.02}
    \begin{tabular*}{\columnwidth}{
        @{\extracolsep{\fill}}
        llccccc
        @{}
    }
        \toprule
        \multirow{2}{*}{Dataset}
        & \multirow{2}{*}{Method}
        & \multicolumn{5}{c}{OA (\%) across Backbones} \\
        \cmidrule(lr){3-7}
        & & PN++ & DGCNN & PC & PCT & PM \\
        \midrule

        \multirow{3}{*}{MN10}
        & PCC
        & 80.9 & 81.3 & 81.9 & 79.5 & 68.2 \\
        & SADM
        & 78.7 & 83.4 & 76.9 & 68.4 & 71.7 \\
        & \textbf{Ours}
        & \textbf{85.1}
        & \textbf{89.9}
        & \textbf{85.2}
        & \textbf{80.1}
        & \textbf{80.9} \\

        \midrule

        \multirow{3}{*}{MN40}
        & PCC
        & 74.5 & 76.5 & 70.4 & 70.2 & 74.6 \\
        & SADM
        & 74.3 & 78.6 & 64.5 & 75.5 & 76.0 \\
        & \textbf{Ours}
        & \textbf{77.5}
        & \textbf{81.3}
        & \textbf{74.6}
        & \textbf{78.9}
        & \textbf{78.9} \\

        \midrule

        \multirow{3}{*}{SN}
        & PCC
        & 68.8 & 70.9 & 65.9 & 68.9 & 68.4 \\
        & SADM
        & 71.2 & 72.9 & 66.3 & 70.8 & 69.9 \\
        & \textbf{Ours}
        & \textbf{80.8}
        & \textbf{81.2}
        & \textbf{76.4}
        & \textbf{78.4}
        & \textbf{80.2} \\

        \midrule

        \multirow{3}{*}{SONN}
        & PCC
        & 27.1 & 27.8 & 31.4 & 21.4 & 22.9 \\
        & SADM
        & 39.3 & 41.3 & 38.3 & 33.0 & 31.3 \\
        & \textbf{Ours}
        & \textbf{46.0}
        & \textbf{49.8}
        & \textbf{43.6}
        & \textbf{39.7}
        & \textbf{40.8} \\

        \bottomrule
    \end{tabular*}

    \caption{Cross-architecture generalization at PPC $=10$. PN++, PC, PCT, and PM denote PointNet++, PointConv, Point Cloud Transformer, and PointMLP, respectively.}
    \label{tab:cross_architecture_results}
    \vspace{-10pt}
\end{table}

%% file: tab/image_dataset.tex
\begin{table}[t]
    \centering
    \small
    \begin{tabular*}{\linewidth}
        {@{\extracolsep{\fill}}lcccc@{}}
        \toprule
        \multirow{2}{*}{Method}
        & \multicolumn{2}{c}{CIFAR-10-LT}
        & \multicolumn{2}{c}{CIFAR-100-LT} \\
        \cmidrule(lr){2-3}
        \cmidrule(lr){4-5}
        & IF $=10$
        & IF $=100$
        & IF $=10$
        & IF $=50$ \\
        \midrule

        LTDD
        & 56.0$\pm$0.3
        & 57.4$\pm$0.1
        & 33.6$\pm$0.2
        & 35.5$\pm$0.1 \\

        RLDD
        & 64.9$\pm$0.2
        & 67.3$\pm$0.1
        & 49.1$\pm$0.1
        & 46.1$\pm$0.2 \\
        Ours
        & \textbf{67.9$\pm$0.1}
        & \textbf{78.1$\pm$0.1}
        & \textbf{52.8$\pm$0.1}
        & \textbf{54.9$\pm$0.1} \\

        \bottomrule
    \end{tabular*}

\caption{Weighted accuracy (\%) under different imbalance factors
at 10 images per class (IPC = 10).}
    \label{tab:cifar_lt}
\end{table}

%% file: fig/feature_compare.tex
\begin{figure}[t]
    \centering
    \includegraphics[width=\linewidth]{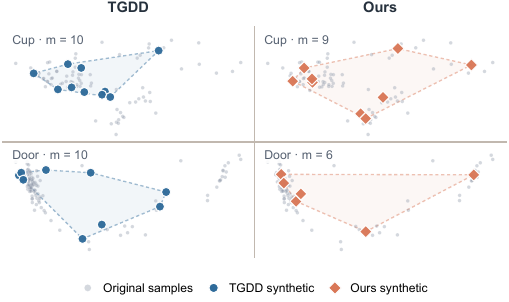}
    \caption{
Feature-space visualization with PCA for original and synthetic tail-class samples. Ours better covers the original distribution with fewer synthetic samples.
    }
    \label{fig:visual}
    \vspace{-10pt}
\end{figure}

%% file: 07_ablation_study.tex
\subsection{Ablation and Further Analysis} 

\subsubsection{Component Ablation.}

\input{tab/each_component}

\input{tab/fine_grained}

\input{fig/initial}
\input{tab/efficiency}

We evaluate each component by removing ASB, GLFA, and PAS individually. As shown in Table~\ref{tab:abeach}, removing any component reduces performance, while the full model performs best on all datasets. In particular, removing ASB causes a 4.1-point drop on ShapeNet55, highlighting the importance of adaptive budget allocation under severe class imbalance. Removing PAS also leads to drops of 1.1--2.3 points across all datasets, demonstrating the benefit of prior-aware supervision. The degradation without GLFA verifies the effectiveness of global-local feature alignment.

\subsubsection{Performance on Head, Middle, and Tail Classes.}

We further divide classes into head, middle, and tail groups based on their training frequencies. As shown in Table~\ref{tab:group_acc}, our method improves head- and middle-class accuracy by 2.2--4.5 points while maintaining competitive tail-class performance. The overall accuracy increases by 2.6 and 3.0 points on ModelNet10 and ModelNet40, respectively, demonstrating that our method improves performance under the original long-tailed distribution rather than simply favoring tail classes.

\subsubsection{Robustness to Initialization.}

We compare four initialization strategies: K-Center~\cite{kcenter}, Herding~\cite{herding}, K-means, and Random. As shown in Fig.~\ref{fig:initialization}, they show noticeable differences at the beginning of optimization, but all improve steadily and reach similar final accuracy. This demonstrates that our method is robust to initialization and can effectively refine different initial synthetic sets. We use K-means as the default initialization.


\subsubsection{Computational Efficiency.}

We compare the computational efficiency of different methods on ModelNet40 at PPC $=10$. As shown in Table~\ref{tab:efficiency}, our method completes synthetic-set optimization in only 31.2 minutes, faster than all competing methods. Including expert training, its total runtime remains comparable to SADM and much lower than PCC and LTDD. It also uses only 6.2 GB of peak GPU memory, demonstrating strong computational efficiency.

%% file: tab/each_component.tex

\begin{table}[t]
\centering
\small
\begin{tabular*}{\columnwidth}{
    @{\extracolsep{\fill}}
    lcccc
    @{}
}
\toprule
\multirow{2}{*}{Variant}
& \multicolumn{4}{c}{Classification Accuracy (\%) } \\
\cmidrule(lr){2-5}
& MN10 & MN40 & SN & SONN \\
\midrule
w/o ASB  & 90.8$\pm$1.2 & 83.2$\pm$0.3 & 79.0$\pm$0.4 & 46.2$\pm$1.2 \\
w/o GLFA & 89.9$\pm$0.8 & 82.3$\pm$0.2 & 81.6$\pm$0.3 & 45.3$\pm$1.7 \\
w/o PAS  & 90.4$\pm$1.1 & 82.9$\pm$0.5 & 81.6$\pm$0.1 & 46.3$\pm$1.2 \\
\textbf{Full}
& \textbf{91.5$\pm$0.5}
& \textbf{84.2$\pm$0.4}
& \textbf{83.1$\pm$0.2}
& \textbf{48.6$\pm$1.2} \\
\bottomrule
\end{tabular*}
\caption{Component ablation at PPC $=10$.}
\label{tab:abeach}
\end{table}

%% file: tab/fine_grained.tex

\begin{table}[t]
    \centering
    \small
    \renewcommand{\arraystretch}{1.02}
    \begin{tabular*}{0.96\columnwidth}{
        @{\extracolsep{\fill}}
        llcccc
        @{}
    }
        \toprule
        \multirow{2}{*}{Dataset}
        & \multirow{2}{*}{Method}
        & \multicolumn{3}{c}{CA (30\%/40\%/30\%)}
        & \multirow{2}{*}{OA} \\
        \cmidrule(lr){3-5}
        & & Head & Middle & Tail & \\
        \midrule
        \multirow{2}{*}{MN10}
        & TGDD
        & 94.0
        & 86.1
        & \textbf{86.2}
        & 88.9 \\
        
        & \textbf{Ours}
        & \textbf{96.2}
        & \textbf{90.6}
        & 85.5
        & \textbf{91.5} \\
        \midrule
        
        \multirow{2}{*}{MN40}
        & TGDD
        & 87.0
        & 72.3
        & \textbf{70.4}
        & 81.2 \\
        
        & \textbf{Ours}
        & \textbf{90.4}
        & \textbf{75.9}
        & 66.0
        & \textbf{84.2} \\
        \bottomrule
    \end{tabular*}
    \caption{Class-wise accuracy (CA, \%) on MN10 and MN40.}
    \label{tab:group_acc}
    \vspace{-10pt}
\end{table}

%% file: fig/initial.tex
\begin{figure}[t]
    \centering
    \includegraphics[width=\linewidth]{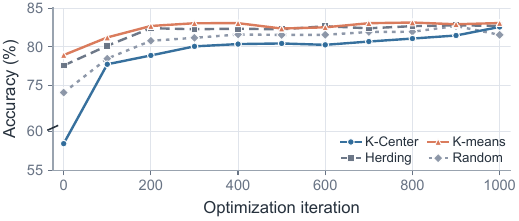}
    \vspace{-10pt}
    \caption{
    Accuracy curves under different initialization strategies during optimization on ShapeNet55 at PPC $=10$, with all strategies converging to comparable final accuracy.
    }
    \label{fig:initialization}
\end{figure}

%% file: tab/efficiency.tex
\begin{table}[t]
    \centering
    \small
    \renewcommand{\arraystretch}{1.02}
    \begin{tabular*}{\columnwidth}{
        @{\extracolsep{\fill}}
        lcccc
        @{}
    }
        \toprule
        \multirow{2}{*}{Method}
        & \multicolumn{3}{c}{Time (min)~$\downarrow$}
        & \multirow{2}{*}{\shortstack{Peak GPU\\Mem. (GB)~$\downarrow$}} \\
        \cmidrule(lr){2-4}
        & Optim. & Expert Train. & Total & \\
        \midrule
        PCC
        & 74.2
        & --
        & 74.2
        & 17.9 \\

        SADM
        & \underline{50.1}
        & --
        & \textbf{50.1}
        & \underline{14.8} \\

        LTDD
        & 87.6
        & \underline{225.3}
        & 312.9
        & 22.4 \\

        \textbf{Ours}
        & \textbf{31.2}
        & \textbf{21.6}
        & \underline{52.8}
        & \textbf{6.2} \\
        \bottomrule
    \end{tabular*}
    \caption{Computational efficiency on ModelNet40 at PPC $=10$ in terms of runtime and peak GPU memory.}
    \label{tab:efficiency}
    \vspace{-10pt}
\end{table}

%% file: 08_conclusion.tex
\section{Conclusion}

We present the first study of long-tailed point cloud dataset distillation, where both the training and test sets can be imbalanced.
We propose a distribution-aware framework that adapts both synthetic budget allocation and expert supervision to the original long-tailed distribution.
This design preserves the utility of long-tailed data under a limited synthetic budget.
Extensive experiments on multiple benchmarks demonstrate consistent improvements under diverse long-tailed settings, highlighting the importance of considering class imbalance in point cloud dataset distillation.
Despite its effectiveness, our method relies on the training distribution for budget allocation and expert supervision, and its performance may degrade under substantial train-test distribution shifts.
Future work will extend this framework to distribution-shifted settings and other imbalanced 3D tasks.